\documentclass[conference]{IEEEtran}
\IEEEoverridecommandlockouts

\usepackage{natbib}
\usepackage{amsmath,amssymb,amsfonts}
\usepackage{algorithmic}
\usepackage{graphicx}
\usepackage{textcomp}
\usepackage{xcolor}
\def\BibTeX{{\rm B\kern-.05em{\sc i\kern-.025em b}\kern-.08em
    T\kern-.1667em\lower.7ex\hbox{E}\kern-.125emX}}

\usepackage{amsthm}
\usepackage{bbold}
\usepackage{subcaption}
\usepackage{algorithm}

\setcitestyle{numbers}
\setcitestyle{square}

\newcommand{\Xb}{\boldsymbol{X}}

\newcommand{\Zb}{\boldsymbol{Z}}

\newcommand{\xb}{\boldsymbol{x}}

\newcommand{\zb}{\boldsymbol{z}}

\newtheorem{theorem}{Theorem}[section]
\newtheorem{proposition}[theorem]{Proposition}
\newtheorem{definition}{Definition}[section]
\DeclareMathOperator*{\argmin}{arg\,min}
\DeclareMathOperator*{\argmax}{arg\,max}

    \author{
   \\
  Department of Statistics\\
  Stanford University\\
  Stanford, CA 94305 \\
  \texttt{roquero@stanford.edu} \\
  \And
  James Zou \\
  Department of Biomedical Data Science \\
  Stanford University \\
  Stanford, CA 94305 \\
  \texttt{jamesz@stanford.edu} 
}
    
\begin{document}

\title{A Unified $f$-divergence Framework Generalizing VAE and GAN}

\author{\IEEEauthorblockN{Jaime Roquero Gimenez}
\IEEEauthorblockA{\textit{Department of Statistics} \\
\textit{Stanford University}\\
Stanford, CA 94305 \\
\texttt{roquero@stanford.edu}}
\and
\IEEEauthorblockN{James Zou}
\IEEEauthorblockA{\textit{Department of Biomedical Data Science} \\
\textit{Stanford University}\\
Stanford, CA 94305 \\
\texttt{jamesz@stanford.edu}}
}

\maketitle

\begin{abstract}Developing deep generative models that flexibly incorporate diverse measures of probability distance is an important area of research. Here we develop an unified mathematical framework of $f$-divergence generative model, f-GM, that incorporates both VAE and $f$-GAN, and enables tractable learning with general $f$-divergences. f-GM allows the experimenter to flexibly design the $f$-divergence function  without changing the structure of the networks or the learning procedure. f-GM jointly models three components: a generator, a inference network and a density estimator. Therefore it simultaneously enables sampling, posterior inference of the latent variable as well as evaluation of the likelihood of an arbitrary datum. f-GM belongs to the class of encoder-decoder GANs: our density estimator can be interpreted as playing the role of a discriminator between samples in the joint space of latent code and observed space. We prove that f-GM naturally simplifies to the standard VAE and to $f$-GAN as special cases, and illustrates the connections between different encoder-decoder GAN architectures. f-GM is compatible with general network architecture and optimizer. We leverage it to experimentally explore the effects---e.g. mode collapse and image sharpness---of different choices of $f$-divergence.
\end{abstract}
\section{Introduction}

In the standard setting for learning generative models, we observe $N$ i.i.d. realizations $\Xb_1, \dots, \Xb_N$ in the space $\mathcal{X}$ following the distribution $p^*$ which we want to learn. We define a parametric model of distributions---the generative model---and then optimize for the best parameter for the model that matches $p^*$ through the observed samples. Generative models are often based on a common structure: a latent variable $\Zb$ in a latent space $\mathcal{Z}$ is sampled from some known simple (potentially parametrized) distribution $p_{\theta}^Z(\zb)$, and then a sample $\Xb \in \mathcal{X}$ is generated from some distribution $p_{\theta}^{X|Z}(\xb|\zb)$, leading to an expression for the distribution of the joint couple $p^{XZ}_{\theta}(\xb,\zb)$. There are two main categories of training procedures for generative models: Variational Auto-Encoders (VAEs) \cite{kingma2013auto} and Generative Adversarial Networks (GANs) \cite{goodfellow2014generative}, each training the generative network $p_{\theta}$ with an auxiliary network. 

It is typically not feasible to directly match $p^*$ and the generative model directly, as we only have access to $p^*$ through the samples, and the marginal $p^X_{\theta}$ of the model over $\mathcal{X}$ is usually not accessible. VAEs introduce a family of distributions $q_{\phi}(\zb|\xb)$---the variational family---and simultaneously optimize generative parameters $\theta$ and the variational parameters $\phi$ so that the distributions $p_{\theta}^{XZ}$ and $(p^* \ltimes q_{\phi})(\xb,\zb) := p^*(\xb)q_{\phi}(\zb|\xb)$ defined over the joint space $\mathcal{X}\times \mathcal{Z}$ are matched. GANs introduce a discriminative network $T_{\lambda}$, a real-valued mapping over $\mathcal{X}$ that is evaluated on the original and generated samples and thus creates a proxy for a loss that is then minimized with respect to the parameters of the generator. A more general class of generative models has been developed by merging these two families: these are called encoder-decoder GANs \cite{donahue2016adversarial,dumoulin2016adversarially}. As in the VAE setting, these models match directly the joint distributions $p_{\theta}^{XZ}$ and $(p^* \ltimes q_{\phi})(\xb,\zb)$. However, instead of deriving parameter updates by gradient descent from a closed form expression of the ``discrepancy'' between these distributions, samples are generated by the generative model and variational model, and then a discriminative network is trained to distinguish between the two samples. Theoretical work has focused on the advantages/drawbacks of these models compared to simpler architectures \cite{arora2018gans}. One contribution of our work is to clarify the relationship between the different members of the encoder-decoder GANs, and with respect to the original VAE and GAN models.

Both VAE and GAN families of models require a measure of discrepancy between probability distributions. A commonly used family of discrepancy is the f-divergence family, many types of VAE and GANs are designed to minimize different such $f$-divergences between the distributions associated with the real samples and the generated ones. Different choices of functions $f$ lead to different outcomes when training a model. In particular, the VAE in \cite{kingma2013auto} uses the Kullback-Leibler divergence, and the GAN in \cite{goodfellow2014generative} uses a slightly transformed version of the Jensen-Shannon divergence. Therefore it is important to have training procedures that allow the scientist to choose the appropriate divergence to the desired task. Models based on some choice of $f$ will tend to suffer from mode collapse \cite{metz2016unrolled}, where the generator outputs samples in a limited subset of the whole initial dataset (i.e. the generator focuses its mass in a mode of the true distribution). Other choices are suspected to lead to blurry outputs when generating images. In consequence, an immediate generalization of the GAN objective through the use of a general f-divergence \cite{nowozin2016f} allows for such flexibility in the GAN family. 

\paragraph{Our Contributions}  This paper develops an unified mathematical framework that incorporates both VAE and GAN and enables tractable learning with general $f$-divergence.   Our $f$ generative model framework, f-GM, allows the experimenter to flexibly choose the $f$-divergence function that suits best the context, without changing the structure of the networks or the learning procedure, which was not feasible previously for encoder-decoder GAN. f-GM jointly models three components: a generator, a inference network and a density estimator. Therefore it simultaneously enables sampling, posterior inference of the latent variable as well as evaluation of the likelihood of an arbitrary sample. We prove that f-GM naturally reduces to the standard VAE and to $f$-GAN as special cases. f-GM is compatible with general network architecture and optimizator, and we leverage it to experimentally explore the effects of different choices of $f$-divergence.

\section{The Unified f-GM Model}\label{section:model}

\paragraph{f-Divergence Definitions And Notations}

We start by defining the f-divergence and the Fenchel conjugate, and providing classical examples of such measures.

\begin{definition}
We say that a mapping $f:[0,+\infty) \rightarrow \mathbb{R}$ is an f-divergence function if $f$ is a proper continuous convex function such that $f(1) = 0$. For any given f-divergence function $f$, we define the associated f-divergence between two probability distributions $p$ and $q$ by $D_f(p||q) = \mathbb{E}_{\Xb \sim q}\big[f\big(\frac{p(\Xb)}{q(\Xb)}\big)\big]$. 

For a f-divergence function $f$, define the Fenchel conjugate $f^*$ by:
\[
f^*(u) = \sup_{x \geq 0} \{ux - f(x)\}
\]
We denote its domain by $D_{f^*} \subset \mathbb{R}$.
\end{definition}
The Fenchel conjugate satisfies the Fenchel-Young inequality, which gives a variational formulation of the initial f-divergence function $f$.

\paragraph{The f-GM Model}
To motivate our general model, we first define the $f$-divergence Variational Auto-encoder ($f$-VAE) as the training procedure which minimizes the $f$-divergence between the joint distributions $p_{\theta}^{XZ}$ and $p^*\ltimes q_{\phi}$. We denote such optimization objective by $L^{V}_f$:
\begin{align*}
L^{V}_f(\theta, \phi) = & D_f(p^*\ltimes q_{\phi}||p_{\theta}^{XZ}) 
\\ = & \mathbb{E}_{(\Xb,\Zb) \sim p_{\theta}^{XZ}}\Big[f\big(\frac{p^*(\Xb)q_{\phi}(\Zb|\Xb)}{p_{\theta}^{XZ}(\Xb,\Zb)}\big)\Big] 
\end{align*}
The usual VAE (that we call KL-VAE) is a particular case of the $f$-VAE, where the $f$-divergence function is $f_{KL}(x) = x\log(x)$. This formulation of $f$-VAE is therefore a natural extension to the KL-VAE which provides flexibility in the choice of the f-divergence function. 

The goal of the $f$-VAE is to find an optimal set of parameters $\theta^*, \eta^*$ such that:
\[
L^{V*}_f := L^V_f(\theta^*, \eta^*) = \inf_{\theta,\eta}L^V_f(\theta,\eta)
\]
Unfortunately, this objective is hard to optimize because approximating the expectation defining $L^V_f$ by a Monte-Carlo average requires evaluating $p^*(\xb)$, which is unknown. Recent work has tried to overcome this issue by adding noise to the samples in $\mathcal{X}$ \cite{zhang2018training}. In order to get around this issue without introducing extra noise, we propose a new variational form of this $f$-divergence. In addition to the generative model $p_{\theta}$ and the variational family $q_{\phi}$, we introduce a \emph{density estimation} model $p_{\eta}$, which encodes a family of distributions over the space $\mathcal{X}$. We now define our novel optimization objective.
\begin{definition}
For a given set of parameters $\theta, \phi, \eta$, define the unified f-divergence optimization objective $L^M_f$ as follows, where $f^*$ is the Fenchel conjugate function of $f$:
\begin{align*}
L^M_f(\theta&,\phi,\eta) = \mathbb{E}_{(\Xb,\Zb)\sim p^* \ltimes q_{\phi}}\Big[f'\left(\frac{p_{\eta}(\Xb)q_{\phi}(\Zb|\Xb)}{p_{\theta}^{XZ}(\Xb,\Zb)}\right)\Big] 
\\ & - \mathbb{E}_{(\Xb,\Zb)\sim p_{\theta}^{XZ}}\Big[f^*\left(f'\left(\frac{p_{\eta}(\Xb)q_{\phi}(\Zb|\Xb)}{p_{\theta}^{XZ}(\Xb,\Zb)}\!\right)\right)\Big]
\end{align*}
\end{definition}
Here $p_{\eta}$ estimates density, $q_{\phi}$ performs posterior inference and $p_{\theta}^{XZ}$ is the generative model. The motivation behind using $L^M_f(\theta,\phi,\eta)$ as a proxy in order to minimize $L^V_f$ is given by the following proposition.
\begin{proposition}\label{prop:VgeqM}
For any set of parameters $\theta, \phi, \eta$, we have the following Fenchel-Young based inequality:
\[
L^M_f(\theta,\phi,\eta) \leq L^V_f(\theta, \phi) = D_f(p^*\ltimes q_{\phi}||p_{\theta}^{XZ})
\]
Furthermore, assuming that $p_{\eta}(\xb)$ can represent any $\mathbb{R}_+$-valued mapping across the choice of $\eta$ (i.e. in the ideal case where density estimator function has ``infinite capacity''), the optimal value $\eta^*$ over the supremum term in $L^*$ is such that $p_{\eta^*} = p^*$, so that we have:
\[
\sup_{\eta} L^M_f(\theta,\phi,\eta) = L^M_f(\theta, \phi, \eta^*) = L^V_f(\theta,\phi)
\]
\end{proposition}
The tool used to derive this inequality---Fenchel-Young---is the same as in the f-GAN work \cite{nowozin2016f}, and originates as a variational characterization of a f-divergence \cite{nguyen2010estimating}. We will later on describe the connection between f-GAN and our model. The crucial difference between our f-GM and previous encoder-decoder GAN architecture is that our model uses Fenchel-Young more ``sparingly'' by replacing the only unknown term in the target expression (that is, $p^*$) through the variational bound, instead of the whole expression. The optimal $\eta^*$ may not exist in practice, the crucial point is that $L^M_f$ is a lower bound of $L^V_f$, and therefore we optimize the quantity $L^M_f(\theta,\phi,\eta)$ to target the optimal value $L^{M*}_f$ given by:
\[
L^{M*}_f = \inf_{\theta, \phi} \sup_{\eta} L^M_f(\theta, \phi, \eta)
\]
We can now present the algorithm f-GM  to train the generative model under the objective $L_{f}^M$. As we approximate $L^M_f$ by an average,  batch size $K$ is a hyperparameter. 
\begin{algorithm}[tb]
\begin{algorithmic}
    \STATE {\bfseries Input:} Dataset $\{\Xb^{data}_i\}_i$, initialized networks $p_{\theta}^{XZ}$, $q_{\phi}$, $p_{\eta}$, number of iterations $T$ and batch size $K$
    \FOR{$t = 1$ {\bfseries to} $T$}
    \FOR{$k = 1$ {\bfseries to} $K$}
    \STATE Sample $\Xb^{data}_k$ from the initial dataset.
    \STATE Compute $q_{\phi}(\cdot|\Xb^{data}_k)$ variational distributions, sample $\Zb^{data}_k\sim q_{\phi}(\cdot |\Xb^{data}_k)$ 
    \STATE Sample $\Zb^{gen}_k \sim p^Z_{\theta}$.
    \STATE Compute $p_{\theta}^{X|Z}(\cdot|\Zb^{gen}_k)$ generative distribution, sample $\Xb^{gen}_k \sim p^{X|Z}_{\theta}(\cdot|\Zb^{gen}_k)$
    \ENDFOR
    \STATE Compute Monte-Carlo approximation: $\hat{L}^M_f(\theta,\phi,\eta) = \sum_{k}f'(\frac{p_{\eta}(\Xb^{data}_k)q_{\phi}(\Zb^{data}_k|\Xb^{data}_k)}{p_{\theta}^{XZ}(\Xb^{data}_k,\Zb^{data}_k)}) - \sum_k f^*(f'(\frac{p_{\eta}(\Xb^{gen}_k)q_{\phi}(\Zb^{gen}_k|\Xb^{gen}_k)}{p_{\theta}^{XZ}(\Xb^{gen}_k,\Zb^{gen}_k)}))$
    \STATE Update: 
    $\begin{cases} 
    \theta \leftarrow \theta - \nabla_{\theta}\hat{L}^M_f(\theta,\phi,\eta)
    \\ \phi \leftarrow \phi - \nabla_{\phi}\hat{L}^M_f(\theta,\phi,\eta)
    \\ \eta \leftarrow \eta + \nabla_{\eta}\hat{L}^M_f(\theta,\phi,\eta)
    \end{cases}
    $
    \ENDFOR
    \STATE $p_{\theta}^{XZ}$, $q_{\phi}$, $p_{\eta}$
    \caption{Unified $f$-divergence generative model algorithm f-GM}\label{algorithm:unified}
\end{algorithmic}
\end{algorithm}

We present a pseudo-code of our unified model in Algorithm~\ref{algorithm:unified}. The main input and output of the algorithm is the model itself, represented by the three networks. For $T$ iterations, a batch of $K$ pairs of samples (one pair based on the real data and variational latent code, the other originating from the generative model) is computed and transformed into the Monte-Carlo approximation of our quantity of interest $L^M_{f}$. Then parameters can be updated following standard optimizer such as Adam \cite{kingma2014adam}. As Proposition \ref{prop:VgeqM} indicates, the optimal value $\eta^*$ is independent of $\theta,\phi$. That is, during training, regardless of the current values of $\theta,\phi$ defining the generative and variational networks, the updates of $\eta$ are such that the gradient steps are taken towards the true desired optimum, and does not depend on the generative network which may not be properly trained.

Our algorithm requires an additional property: $q_{\phi}$ and $p_{\theta}^{X|Z}$ are chosen such that we can compute the gradients with respect to the parameters $\theta, \phi$ in the Monte-Carlo approximation of $L^M_f(\theta,\phi,\eta)$. This is a standard property due to the dependence in the parameters of the samples as they are generated. We assume that the generative model $p_{\theta}^{X|Z}$ and the variational family $q_{\phi}$ are designed using methods involving normalizing flows \cite{dinh2016density, rezende2015variational} or other variants of the reparametrization trick \cite{rezende2014stochastic, kingma2013auto} and allow for such gradient computations.

Setting aside issues related to the training process, the optimal solution to the optimization problem is exactly what we hope for and is identifiable.
\begin{proposition}\label{Optimality}
Assuming that $p_{\eta}(\xb)$, $q_{\phi}(\zb|\xb)$ and $p_{\theta}(\xb|\zb)$ can represent any $\mathbb{R}_+$-valued mapping (i.e. in the ideal case where networks have ``infinite capacity''), the optimal values $\eta^*, \phi^*, \theta^*$ are such that $p_{\eta^*} = \int_{\mathcal{Z}}p(\zb)p_{\theta^*}(\xb|\zb)d\zb = p^*$ and $q_{\phi^*}(\zb|\xb) = p_{\theta^*}(\zb|\xb)$.
\end{proposition}
All three networks of our new objective are useful by themselves. $L^M_f$ has the nice property that all the networks introduced are of interest and not just mere auxiliary tools to help with the training process. When the optimization in f-GM works well, the three corresponding trained networks solve each of the following three important problems: 
\begin{itemize}
\setlength\itemsep{-0.2em}
\item \textbf{Parameter Estimation / Sampling}: The generative network $p_{\theta}^{XZ}(\xb,\zb)$ is such that we can generate samples from the true distribution $p^*$ by ancestral sampling. The marginal over $\mathcal{X}$ of the generative distribution $p_{\theta}^X(\xb) = \sum_{\zb}p_{\theta}^{XZ}(\xb,\zb)$ is equal to $p^*(\xb)$.
\item \textbf{Inference}: The variational network $q_{\phi}(\zb|\xb)$ recovers a latent code for a given observed sample, which corresponds to  $p_{\theta}^{Z|X}(\zb|\xb) = p_{\theta}^{XZ}(\xb,\zb)/p_{\theta}^X(\xb)$ (posterior generative distribution).
\item \textbf{Density Estimation}: The density estimator network $p_{\eta}$ approximates the true density function $p^*$. It can evaluate the likelihood of an arbitrary point and complements the generative network.
\end{itemize}

Other models are also able to solve several of these goals, such as reversible generative models \cite{kingma2018glow}. Flow networks that directly map the latent space $\mathcal{Z}$ to $\mathcal{X}$ through an invertible function also solve the above problems \cite{dinh2014nice,rezende2015variational, dinh2016density}. However, all these models can not be applied to generative models where we know the generative process has a given pre-defined structure based on prior biological or physical knowledge.

\section{Recovering Previous f-divergence Generative Models from f-GM}\label{section:reduction}

We now show how our model encapsulates the two main families of $f$-divergence based generative models. The following simplifications are not based on interpolations of optimization objectives nor on concatenation of generative models on top of another as in \cite{mescheder2017adversarial, larsen2015autoencoding, chen2016infogan}. Our model does not generalize other generative models that take into account the geometry of the space $\mathcal{X}$, such as models based on an optimal transport distance \cite{arjovsky2017wasserstein}. 

\paragraph{Simplification Into Auto-encoder: The KL-divergence Case And Fenchel-Young Equality}
As previously mentioned, the KL-VAE is a particular case of the $f$-VAE. Whenever we choose $f_{KL}$ as the $f$-divergence function, our objective $L^M_{f_{KL}}$ simplifies directly into an optimization objective equivalent to that of the KL-VAE.
\begin{proposition}\label{MeqVKL}
Given our objective $L^M_{f}(\theta, \phi,\eta)$ and the KL-VAE objective $L^V_{f_{KL}}(\theta,\phi)$, we have the following identity:
\[
L^M_{f_{KL}}(\theta,\phi,\eta) = L^{V}_{f_{KL}}(\theta, \phi) - KL(p^*||p_{\eta})
\]
\end{proposition}
Our $L^M_{f_{KL}}$ introduces a new network $p_{\eta}$, thus becoming a lower bound of the KL-VAE target $L^V_{f_{KL}}$. This proposition shows that for the specific choice of $f_{KL}$, the step that introduces $p_{\eta}$ can be ignored. The update steps over $\theta, \phi$ are the same ones as in KL-VAE, the terms depending on $\eta$ and those depending on $\theta, \phi$ decouple. This suggests a fundamental reason why one can train the KL-VAE objective but it's hard to directly optimize $f$-VAE. If we expand the expression for $L^V_{f_{KL}}$ we get: 
\begin{align*}\label{equation:kl-vae}
L^{V}_{f_{KL}}&(\theta, \eta) =  \text{KL}(p^*\ltimes q_{\phi}||p_{\theta}) 
\\ = & \mathbb{E}_{(\Xb, \Zb)\sim p^*\ltimes q_{\phi}}\left[\log \frac{p^*\ltimes q_{\phi}(\Xb,\Zb)}{p_{\theta}(\Xb,\Zb)}\right] 
\\ = & \mathbb{E}_{\Xb \sim p^*}\!\left[\log p^*(\Xb)\right] \!-\! \mathbb{E}_{(\Xb, \Zb)\sim p^*\ltimes q_{\phi}}\!\left[\log \frac{p_{\theta}(\Xb,\Zb)}{q_{\phi}(\Zb|\Xb)}\!\right]
\end{align*}
For this particular choice of divergence, the evaluation of $p^*$ is again decoupled from the terms containing the parameters, therefore the usual VAE $L^V_{f_{KL}}$ can be trained, unlike the general form $L^V_{f}$.

For general $f$-divergence functions, our model simplifies into the $f$-VAE whenever the density estimator attains the optimal value in the Fenchel-Young inequality, as shown in Proposition~\ref{prop:VgeqM}. Therefore, $f$-VAE is encapsulated in our model, as optimizing $L^V_f(\theta,\phi)$ is exactly equivalent to optimizing our model $L^M_f(\theta,\phi,\eta^*)$ under a perfect choice of parameter values for the density estimator network. 

\begin{figure*}[ht!]
\centering
\includegraphics[width=0.8\linewidth]{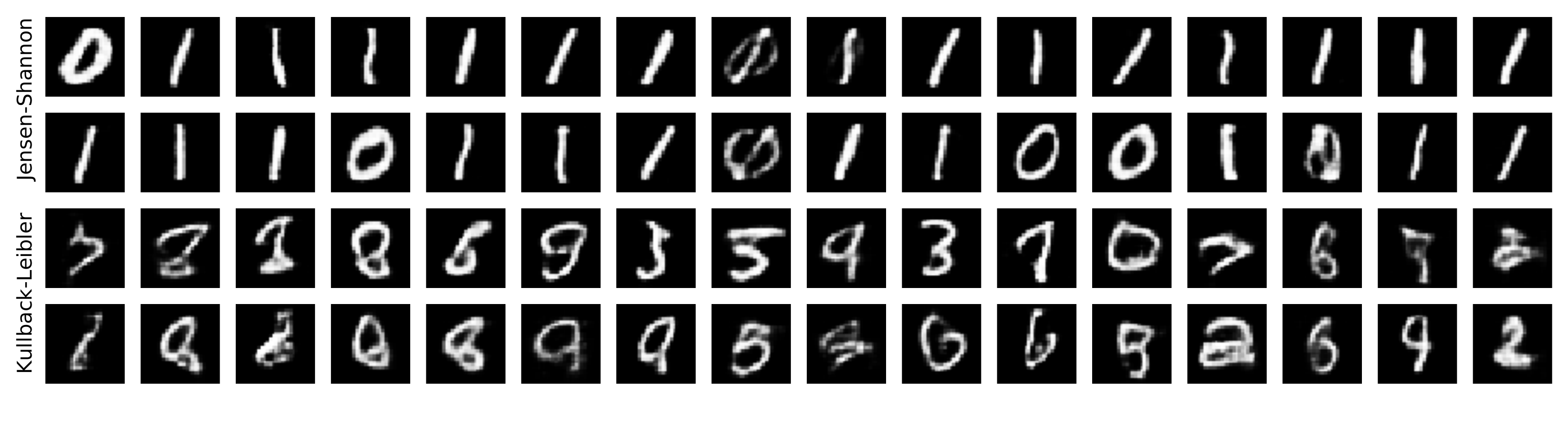}
\end{figure*}
\begin{figure*}[ht!]
\centering
\includegraphics[width=0.8\linewidth]{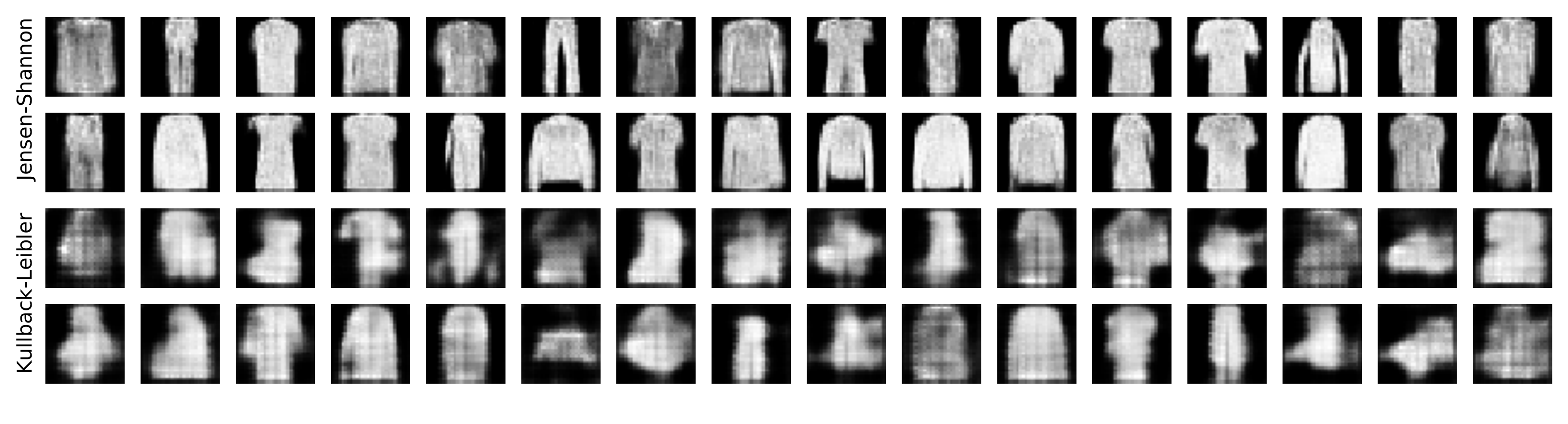}
\caption{\textbf{Comparison on MNIST (top) and Fashion-MNIST (bottom) between f-GM with Kullback-Leibler and Jensen-Shannon divergences.} First two rows of each block show a random selection of outputs from a generator trained with our model and the Jensen-Shannon divergence. Bottom two rows show the outputs when our model is trained the Kullback-Leibler divergence.}
\label{fig:fashion}
\end{figure*}

\paragraph{Simplification Into Adversarial Networks}
Our f-GM and the $f$-GAN \cite{nowozin2016f} both use a network to replace an intractable or unknown term---$p^*$ for us or the likelihood ratio for $f$-GAN---through a variational expression. Our model contains $f$-GAN model as a particular case, which we show next. Instead of considering the $f$-divergence in the joint model $D_f(p^*\ltimes q_{\phi}||p_{\theta}^{XZ})$, the $f$-GAN wants to minimize the $f$-divergence between the marginal distributions $D_f(p^*||p_{\theta}^X)$. This expression is defined by an expectation in terms of $p^*, p_{\theta}^X$, which can not be evaluated. The f-GAN derives a lower bound based on the same Fenchel-Young inequality as in our model, that is subsequently maximized to give a proxy of the target $D_f(p^*||p_{\theta}^X)$. Such proxy is then minimized over the generative parameters. Introducing a \emph{discriminator} network $T_{\lambda}(\xb)$, let $L^G_f(\theta,\lambda)$ be the $f$-GAN optimization objective defined as follows and satisfying the following inequality:
\begin{align}
L^G_f(\theta,\lambda)  = & \mathbb{E}_{\Xb\sim p^*}\big[T_{\lambda}(\Xb)\big] - \mathbb{E}_{\Xb \sim p_{\theta}^X}\big[f^*(T_{\lambda}(\Xb))\big] \nonumber
\\ \leq & \mathbb{E}_{\Xb \sim p_{\theta}^X}\Big[f\left(\frac{p^*(\Xb)}{p_{\theta}^X(\Xb)}\right)\Big] = D_f(p^*||p_{\theta}^X) 
\end{align}
The $f$-GAN targets the optimal value $L^{G*}_f = \inf_{\theta}\sup_{\lambda}L^G_f(\theta, \lambda)$, where the $L^G_f(\theta,\lambda)$ is a tractable lower bound of $D_f(p^*||p_{\theta}^X)$, which is first maximized over $\lambda$. We now see how our model simplifies into the $f$-GAN. Assuming that for any given generative network $p_{\theta}$ there is an optimal value $\phi^*$ such that the corresponding variational network $q_{\phi^*}$ perfectly matches the posterior distribution $p_{\theta}^{Z|X}(\zb|\xb)$ associated with the generative network (i.e. $q_{\phi}$ has ``infinite capacity''), the following simplification occurs:
\begin{proposition}\label{prop:simplification}
Let $\phi^* = \phi^*(\theta)$ be the optimal value such that $q_{\phi^*}(\zb|\xb) = p_{\theta}^{Z|X}(\zb|\xb)$. Then \[
L^M_f(\theta,\phi^*,\eta) = L^G_f(\theta,\eta)
\]
\end{proposition}

Our model can therefore be simplified into a $f$-GAN by associating the term $f'\big(\frac{p^*(\xb)}{p_{\theta}^X(\xb)}\big)$ to the discriminator $T_{\lambda}$. Note that our assumption about $\phi^*$ is equivalent to assuming that we have access to the posterior generative distribution (and, equivalently, to the marginal $p_{\theta}^X$). The purpose of these derivations is to show the theoretical equivalence of solving the f-GAN problem and solving ours under the ideal choice of parameters, to prove how our model encapsulates f-GAN. There is one important difference with respect to the previous paragraph: given that we can not swap the terms $\inf_{\phi}$ and $\sup_{\eta}$ in the optimization objective of $L^M_{f}(\theta,\phi,\eta)$, we get that the optimal $\phi^*$ for which the previous simplification of our model into f-VAE is not $\argmin_{\phi} L^M_{f}(\theta,\phi,\eta)$. We do not attain such $\phi^*$ by freezing the other parameters and minimizing over $\phi$.

\begin{figure}[t]
\centering
\begin{subfigure}[b]{0.4\textwidth}
\includegraphics[width=\linewidth]{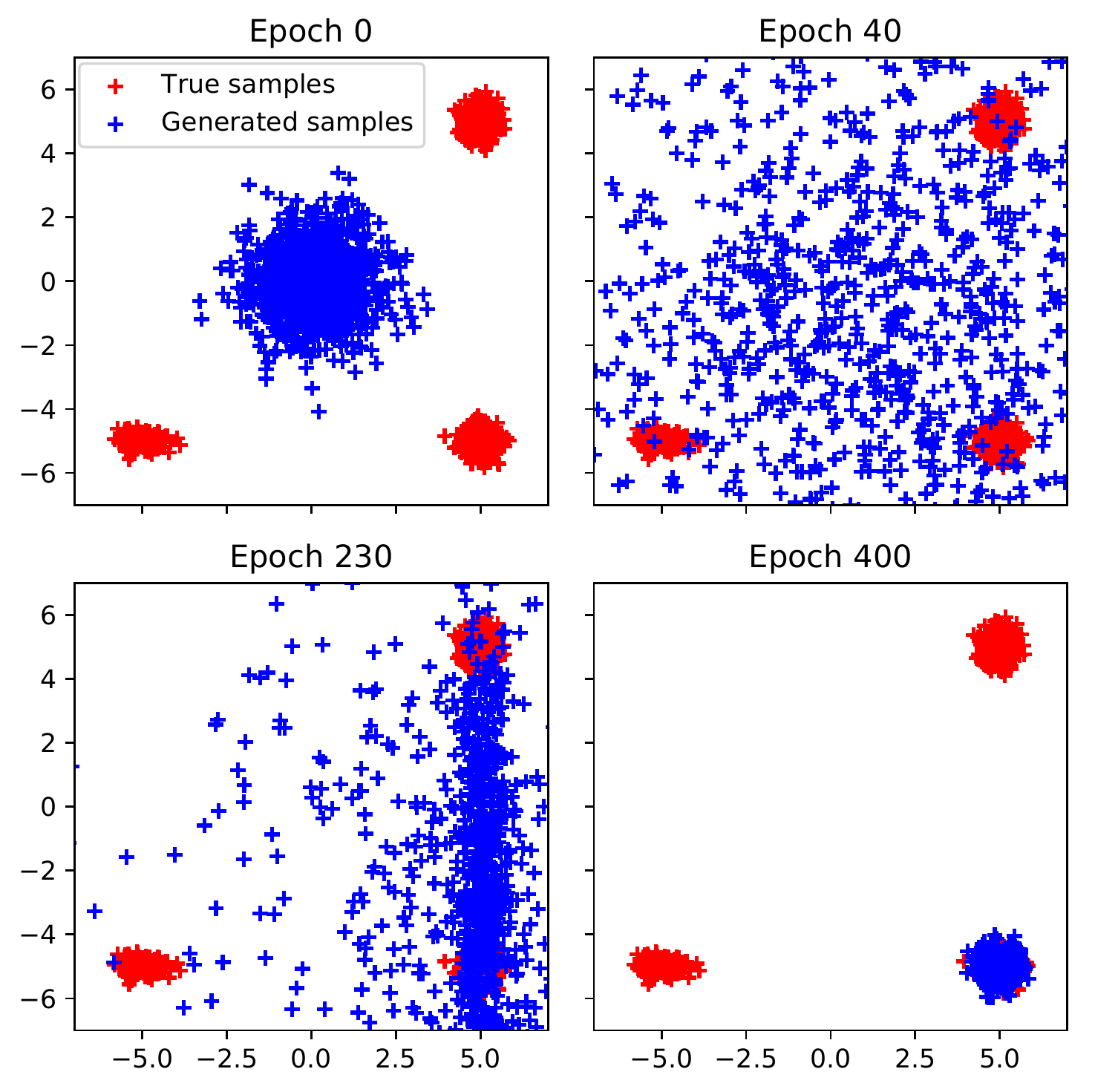}
\end{subfigure}
\caption{Samples from true distribution (red) and from generated distribution (blue) at four different epochs, showing mode collapse of the generator when trained with Jensen-Shannon.}
\label{fig:diagnostics}
\end{figure}

\section{Experiments}\label{section:experiments}

\paragraph{Experiments On MNIST And Fashion-MNIST}

We implemented f-GM and fit it with two different $f$-divergence functions to illustrate the flexibility of our model and the importance of an appropriate choice of $f$. We run experiments to compare the Kullback-Leibler divergence and the Jensen-Shannon divergence. GANs based on Jensen-Shannon (JS) divergence are known to suffer mode collapse, whereas the VAE, based on Kullback-Leibler (KL), usually outputs noisy images. We validate it with experiments on MNIST and Fashion-MNIST \cite{lecun2010mnist, xiao2017fashion}.

For MNIST, f-GM with JS outputs only two distinct sharp digits, in a clear case of mode collapse (Top two rows in upper Figure~\ref{fig:fashion}). Similarly, for Fashion-MNIST, the JS version of f-GM outputs two or three different (among 10) types of clothing, with very sharp images (Top two rows in lower Figure~\ref{fig:fashion}).
Conversely, when using f-GM with KL, all the elements of the dataset are represented in the generated samples. However, the digits in MNIST are fuzzier, as well as the clothes images in Fashion-MNIST (bottom two rows in upper and lower Figure~\ref{fig:fashion} respectively). While similar phenomenon has been discussed in other models, f-GM makes it easy to explore the trade-offs from different choices of $f$ in the same framework.

\paragraph{Diagnosing Mode Collapse Through The Evaluation Of The Density Estimator On The Real Dataset}

f-GM has the nice capability of self-diagnosing during training in order to detect mode collapse. We illustrate this with an example based on mode collapse on a mixture of Gaussians, which has received widespread attention \cite{srivastava2017veegan, metz2016unrolled, lin2018pacgan}. By using the Jensen-Shannon divergence in our model, at some point during training on a well separated mixture of Gaussians the generator focuses on one mode of the true distribution (Figure~\ref{fig:diagnostics}(b)).

\section{Comparison to Related Works and Discussion}\label{section:discussion}

\paragraph{Comparison To Related Works}

f-GM has several key differences and advantages compared to the existing deep generative models. First, by Prop.~\ref{prop:VgeqM}, the objective of our method is to \emph{match the joint models} $p_{\theta}^{XZ}$ and $p^* \ltimes q_{\phi}$ in $f$-divergence. We are not matching marginals only in the $\mathcal{X}$ space as the $f$-GAN does, nor we are using different losses for different sections of the model \cite{makhzani2015adversarial, pu2017adversarial, zhao2018information}. Similar to the BiGAN/ALI models \cite{dumoulin2016adversarially, donahue2016adversarial}, we use the Fenchel-Young inequality to approximate the $f$-divergence over the joint distributions, but instead of substituting the whole likelihood ratio by an auxiliary term, we replace $p^*$, \emph{the one term we do not have access to}, and keep the terms $q_{\phi}$ and $p_{\theta}^{XZ}$ in the expectation. By reusing the generative and variational models, the new ``discriminative'' network---that is, our density estimator $p_{\eta}$---is only a mapping over $\mathcal{X}$ instead of a mapping over $\mathcal{X}\times \mathcal{Z}$, providing savings in terms of parameters w.r.t. BiGAN/ALI. If we focus on the value at which equality is attained in Fenchel-Young for $f$-GAN, and whenever $\theta = \theta^*$ is such that $p_{\theta^*} = p^*$ ---i.e. perfect training of the generator---we get that $T_{\lambda^*}(\Xb) = f'\big(\frac{p^*(\Xb)}{p_{\theta}^X(\Xb)}\big)$ is a constant. More generally, a adversarial network that targets a density ratio \cite{uehara2016generative} is such that the optimal value of the discriminator is of little value. In contrast, equality is attained in Fenchel-Young in our model whenever $p_{\eta^*} = p^*$, which is of interest by itself as a density estimator. Moreover, the optimal parameter $\eta^*$ in our model \emph{does not depend on the values of the other parameters} $\theta, \phi$. That is, $\argmax_{\eta} L^M_f(\theta,\phi,\eta)$ does not depend on $\theta, \phi$, unlike $\argmax_{\lambda} L^G_f(\theta, \lambda)$ which depends on $\theta$. That means that regardless of the value of $\theta$, optimizing the parameter $\eta$ in our unified model always yields the same result, that is, $p_{\eta^*} = p^*$. Our construction disentangles the generative and variational model from $p_{\eta}$, so that during training the updates in $\eta$ are somehow independent of the other parameters.
\paragraph{Conclusion}

Beyond providing a unified mathematical framework for f-GAN and VAE, which is worthwhile, there are several practical applications of our f-GM framework. First, f-GM allows the researcher to try different divergence functions f, and optimize the model using the same optimization method in Algorithm~\ref{algorithm:unified}. This allows one to directly compare the trade-offs between different choices of f---e.g. the trade-off between mode-collapse and sharpness demonstrated in Figure~\ref{fig:fashion}---which is of great interest to the community. A second application is that the density estimator network of f-GM is useful to detect mode collapse by visualizing how the likelihood of the observed data changes during training.

\bibliographystyle{unsrtnat}

\bibliography{main}

\end{document}